\documentclass[letterpaper]{article} 
\usepackage{aaai2026}  
\usepackage{times}  
\usepackage{helvet}  
\usepackage{courier}  
\usepackage[hyphens]{url}  
\usepackage{graphicx} 
\usepackage{natbib}  
\usepackage{caption} 
\usepackage{algorithm}
\usepackage{algorithmic}

\usepackage{newfloat}
\usepackage{listings}
\DeclareCaptionStyle{ruled}{labelfont=normalfont,labelsep=colon,strut=off} 
\floatstyle{ruled}
\newfloat{listing}{tb}{lst}{}
\floatname{listing}{Listing}
\usepackage{microtype}

\usepackage{amsmath,amssymb,amsthm}
\usepackage{mathtools}
\usepackage{enumitem}
\usepackage{booktabs}
\PassOptionsToPackage{monochrome}{xcolor}
\newcommand{\M}{\mathcal{M}}

\newtheorem{definition}{Definition}

\newtheorem{proposition}{Proposition}
\newtheorem{lemma}{Lemma}
\newtheorem{theorem}{Theorem}
\newtheorem{remark}{Remark}
\newtheorem{corollary}{Corollary}

\newcommand{\C}{\mathcal{C}}

\newcommand{\I}{\mathcal{I}}

\newcommand{\cl}{\operatorname{cl}}

\title{Structured Personalization: Modeling Constraints as Matroids for Data-Minimal LLM Agents}
\author{
    Daniel Platnick\textsuperscript{\rm 1,\rm 2}, Marjan Alirezaie\textsuperscript{\rm 1,\rm 2}, Hossein Rahnama\textsuperscript{\rm 1,\rm 2,\rm 3}
}
\affiliations{
    \textsuperscript{\rm 1}Flybits Labs, Creative Ai Hub\\    
    \textsuperscript{\rm 2}Toronto Metropolitan University\\
    \textsuperscript{\rm 3}MIT Media Lab\\

    \{daniel.platnick,marjan.alirezaie\}@flybits.com, rahnama@mit.edu
}

\begin{document}

\maketitle

\pagestyle{plain}

\begin{abstract}
Personalizing Large Language Model (LLM) agents requires conditioning them on user-specific data, creating a critical trade-off between task utility and data disclosure. While the utility of adding user data often exhibits diminishing returns (i.e., submodularity), enabling near-optimal greedy selection, real-world personalization is complicated by structural constraints. These include logical dependencies (e.g., selecting fact A requires fact B), categorical quotas (e.g., select at most one writing style), and hierarchical rules (e.g., select at most two social media preferences, of which at most one can be for a professional network). These constraints violate the assumptions of standard subset selection algorithms. We propose a principled method to formally model such constraints. We introduce a compilation process that transforms a user's knowledge graph with dependencies into a set of abstract \emph{macro-facets}. Our central result is a proof that common hierarchical and quota-based constraints over these macro-facets form a valid \textbf{laminar matroid}. This theoretical characterization lets us cast structured personalization as submodular maximization under a matroid constraint, enabling greedy with constant-factor guarantees (and $(1\!-\!1/e)$ via continuous greedy) for a much richer and more realistic class of problems.
\end{abstract}


\section{Introduction}
\label{sec:intro}
The effectiveness of LLM-based agents is greatly enhanced by personalization, typically achieved by including user-specific data in the agent's context. However, this creates a fundamental trade-off: adding more data can increase task performance but also inflates token costs and privacy risks associated with data disclosure. This requires a principled approach to \textbf{data minimization}: selecting the smallest possible subset of user data that achieves a desired level of utility.

The utility gained from adding successive pieces of user information often exhibits \textbf{diminishing returns} \cite{Krause-2010-privacy}. 
For instance, providing a first stylistic preference significantly improves an agent's writing, but adding a third or fourth synonymous preference yields far less marginal benefit. 
This property, known as \textbf{submodularity} in discrete optimization, is powerful. 
For utility functions that are submodular, a simple greedy algorithm that iteratively selects the item with the highest marginal utility is provably near-optimal \cite{Nemhauser1978Submodular}.

However, this elegant model breaks down when faced with the structural realities of user data. A user's digital identity is not an unstructured bag of facts. It contains:
\begin{enumerate}[nosep]
    \item \textbf{Logical Dependencies:} Selection necessitates logical completeness. For example, domain-specific entities often rely on internal definitions. Selecting an opaque identifier `(User, lead\_on, ProjectPinnacle)' requires selecting its descriptor `(ProjectPinnacle, is, LegacyMigration)' to prevent the agent from inferring an incorrect strategic context based on the identifier alone. A selection algorithm must respect these dependencies to ensure contextual integrity.
    \item \textbf{Hierarchical Quotas:} A user may impose nested limits. For example, select ``at most 3 hobbies,'' of which ``at most 1 can be a water sport.'' This imposes nested ``at-most-k'' constraints on different categories of data. Simple categorical quotas are a special case of this structure.
    
    Hierarchical quotas can be extended to govern temporal or privacy constraints:
    E.g., Some user information is time-sensitive or restricted by policy, limiting when or how it can be utilized in personalization.
    Additionally, they can encode constraints on the relevance of user information often depends on task or context: E.g., a user's ``preferred writing tone'' may be crucial for email generation but irrelevant for summarization.

\end{enumerate}
These constraints violate the simple independence assumption required by standard greedy algorithms, rendering their performance guarantees void. 
Our work addresses this by formally modeling the complex system of rules as a laminar matroid—a combinatorial structure that captures the essential properties of independence needed to guarantee the performance of greedy algorithms. 
This key reformulation transforms the problem into the well-studied domain of maximizing a submodular function subject to a matroid constraint, for which a simple greedy approach is once again provably near-optimal.
We achieve this by compiling dependencies into macro-facets and enforcing quotas pre-closure.

\paragraph{Contribution.} We demonstrate how to compile a user's identity knowledge graph, complete with logical dependencies, into a set of abstract \emph{macro-facets}, each representing a logically inseparable bundle of attributes that function as a single, atomic unit for selection.
We then prove that applying hierarchical quota constraints to this new set constitutes a \emph{laminar matroid}—a class of matroids that allows constraints to be hierarchically nested in addition to being mutually exclusive \cite{Korte-2012-combinatorial-theory,Schrijver-2003-combinatorial-efficiency}.
This result is significant because it transforms the messy, constrained selection problem into the well-understood problem of maximizing a monotone submodular function subject to a matroid constraint \cite{bilmes2022submodularitymachinelearningartificial}. 
This allows us to retain the provable near-optimality of greedy algorithms while handling a much more expressive and practical class of personalization constraints.

\section{Problem Formulation}
\label{sec:formulation}
We model a user's identity as 
a structured collection of semantically related facts, called \textit{facets}, organized in a knowledge graph.
This graph, denoted as a \textit{chronicle}, evolves as the user interacts with the system.
Intuitively, a chronicle captures what the system knows about the user (e.g., the state of preferences, attributes, and behaviors, at a given point in time).

\begin{definition}[Chronicle and Utility]
Let $\C_u = \{e_1, \dots, e_n\}$ be a finite ground set of facets comprising the user's chronicle. A subset of these facets, $S \subseteq \C_u$, forms the personalization filter used to condition the LLM agent.
The utility of this filter is given by a set function $U: 2^{\C_u} \to \mathbb{R}_{\ge 0}$, which quantifies the agent's performance on a personalization task for a given set of facets. 
\end{definition}
We assume $U$ is \textbf{normalized} ($U(\emptyset)=0$), \textbf{monotone} (i.e., $S \subseteq T \implies U(S) \le U(T)$), and \textbf{submodular}.
This means it exhibits a diminishing returns property, where the utility gained from adding new information is less than or equal to the utility gained from adding that same information to a smaller, existing set.

In practice, $U$ can be estimated empirically by measuring the downstream task performance of the personalized LLM agent (e.g., using an \emph{LLM-as-a-Judge} evaluation framework, over a benchmark suite).

\begin{definition}[Submodularity]
A set function $U$ is submodular if for all subsets $A \subseteq B \subset \C_u$ and any element $e \in \C_u \setminus B$, it satisfies diminishing returns:
$$ U(A \cup \{e\}) - U(A) \ge U(B \cup \{e\}) - U(B). $$

That means the benefit of adding an item $e$ decreases as the existing set $B$ gets bigger. This diminishing-returns property is what allows greedy selection to admit provable approximation guarantees under suitable feasibility constraints.
\end{definition}

\begin{remark}[Justification of Submodularity in LLMs]
The assumption of submodularity is grounded in the principle of \textbf{diminishing returns}. The initial selection of high-salience user data provides the most significant reduction in generation uncertainty. As the context becomes populated, the marginal utility of subsequent facts diminishes due to information redundancy. We define the optimization problem within the effective context window, where the addition of relevant information is strictly beneficial or neutral (monotone).
\end{remark}

The goal is to select a subset $S$ that maximizes $U(S)$ under feasibility constraints.

\paragraph{Logical Dependencies.} We represent dependencies using a directed graph $G = (\C_u, E)$, where an edge $(e_i, e_j) \in E$ means selecting $e_i$ logically entails $e_j$. 
We consider the full set of required facets for any selection $S$ as its \textbf{closure}: the set containing $S$ itself plus all other facets that are transitively implied by its members. 
Any valid personalization filter must be a closed set to be logically complete.

\begin{definition}[Closure]
For any $S \subseteq \C_u$, its closure $\cl_G(S)$ is the set of all facets in $\C_u$ reachable from any facet in $S$ via a path in $G$. Intuitively, $\cl_G(S)$ includes $S$ together with all facets logically or structurally implied by them. 
A selection $S'$ is considered \emph{closed}, or a logically complete subset, if it already contains all its implied facets, i.e., $S' = \cl_G(S')$.
\end{definition}
Any valid personalization filter must be a closed set. While LLMs can infer general semantic entailment, strict closure is essential to safeguard \textbf{contextual integrity} and ensure logical consistency in personal, proprietary, and highly regulated domains. By preventing the disclosure of opaque tokens—such as internal code names or private identifiers—without their defining descriptors, closure precludes hallucinations arising from incomplete or contradictory contexts.
However, enforcing this requirement complicates selection, as adding one facet may force the inclusion of many others.

\paragraph{Hierarchical Quotas.} We often need to limit the number of facets within hierarchically organized categories (e.g., taxonomy of user attributes). 
For instance, if the chronicle contains facets about social media preferences, we might require that a valid selection $S$ includes at most three such preferences in total, and of those three, at most one may concern a professional network (a sub-category). 
This kind of constraint naturally induces a laminar structure over the facet hierarchy.

\section{From Dependencies to Matroids}
\label{sec:theory}
Directly enforcing quota constraints on the raw facets is non-trivial because logical dependencies may exist that cause selecting one facet to implicitly include others, resulting in closure violating the intended quotas.  
Consider a consultant whose profile strictly forbids sharing ``Client Identities'' (Quota $=0$) to avoid conflicts of interest. Selecting a work history item like `(User, led, ProjectApollo)' seems safe. However, if the chronicle defines `(ProjectApollo, client\_is, AcmeCorp)', the closure forces the inclusion of `AcmeCorp', violating the prohibition on client disclosure.
The core of our contribution is to define a new ground set that elegantly absorbs the dependency structure, reframing the problem into a form where greedy selection is once again provably near-optimal.

\subsection{Compiling Dependencies into Macro-Facets}
We first identify logically inseparable bundles of facets using the structure of the implication graph $G$.

\begin{definition}[Macro-Facets]
A \textbf{macro-facet} is a strongly connected component (SCC) of the implication graph $G$. Let $\M$ be the set of all macro-facets. For any selection of macro-facets $S_\M \subseteq \M$, the corresponding set of raw facets is its \textbf{expansion}, defined as the union of their closures:
$$ \operatorname{Exp}(S_\M) \triangleq \bigcup_{m \in S_\M} \cl_G(m) $$
\end{definition}
By this construction, selecting a single macro-facet implicitly selects its entire closure, automatically satisfying all logical dependencies. We can now define the utility function over this new ground set, $U'(S_\M) \triangleq U(\operatorname{Exp}(S_\M))$.

\begin{proposition}
\label{prop:exp-submod}
If the utility function $U$ over $\C_u$ is monotone and submodular, then the utility function $U'$ over $\M$ is also monotone and submodular.
\end{proposition}
Full proofs are provided in the appendix. The key insight is that the expansion operator preserves the necessary set relations for the submodularity inequality to hold.

\subsection{Feasibility as a Laminar Matroid}
With logical dependencies now encapsulated inside macro-facets, the feasible units of selection are closure-consistent. 
We can therefore impose hierarchical quota constraints directly on the macro-facets. This structure forms a laminar matroid.

\begin{definition}[Laminar Family]
A family of subsets $\mathcal{F} \subseteq 2^\M$ is a \textbf{laminar family} if for any two sets $A, B \in \mathcal{F}$, one of the following holds: $A \subseteq B$, $B \subseteq A$, or $A \cap B = \emptyset$.
\end{definition}

\begin{definition}[Laminar Matroid over Macro-Facets]
Let $\mathcal{F}$ be a laminar family on the ground set $\M$. For each set $A \in \mathcal{F}$, let $q_A \in \mathbb{Z}_{\ge 0}$ be a non-negative integer quota. A set of macro-facets $S_\M \subseteq \M$ is defined as \textbf{independent} if for every set $A \in \mathcal{F}$, it satisfies the quota constraint:
$$ |S_\M \cap A| \le q_A. $$
Let $\I$ be the family of all such independent sets.
\end{definition}

\noindent\textbf{Feasibility Convention (Pre-Closure).} All quotas are enforced on the set of \emph{chosen macro-facets} $S_\M$ (before expansion); $\operatorname{Exp}(\cdot)$ affects only the objective (and optional costs), and elements introduced by expansion are \emph{never} counted toward quotas. Mutual exclusivity between items within the same parent category can be expressed by adding a nested subset $\{m_i,m_j\}$ with $q=1$. Quotas apply only to chosen macro-facets because each macro-facet already internalizes all its dependent facets. Counting post-expansion elements would double-count implied information rather than independent choices. (Implementation note: skipping a candidate whose expansion is already covered is a utility shortcut, not a feasibility rule.)

\begin{theorem}
The system $(\M, \I)$ is a matroid by the standard augmentation argument for laminar systems \citep{Schrijver-2003-combinatorial-efficiency}.

\label{thm:laminar-matroid}
\end{theorem}
\begin{proof}[Proof Sketch]
We verify the three matroid axioms. The empty set axiom and downward closure axiom are satisfied by construction. The critical step is proving the augmentation property: for any two independent sets $A, B \in \I$ with $|A| < |B|$, there exists an element $x \in B \setminus A$ such that $A \cup \{x\} \in \I$.

The proof proceeds by contradiction. If no such $x$ exists, then for every $x \in B \setminus A$, adding it to $A$ must violate some quota. This implies that for each such $x$, there is a ``tight'' set in the laminar family $\mathcal{F}$ that contains $x$ and is already at its capacity with elements from $A$. By leveraging the nested or disjoint property of laminar families, we can show that this leads to a contradiction on the relative sizes of $A$ and $B$. Specifically, it would imply $|A| \ge |B|$, which contradicts our initial assumption. Therefore, an element $x$ with the desired property must exist. The full proof, which follows the standard construction of laminar matroids \cite{fisher1978analysis} and is adapted here to the macro-facet setting, can be found in the appendix.
\end{proof}

This laminar-matroid formulation yields special cases such as partition matroids, and algorithmic consequences including standard greedy approximation guarantees.

\begin{corollary}[Partition Matroids as a Special Case]
If the laminar family $\mathcal{F}$ is simply a partition of the macro-facets $\M = \biguplus_{t \in T} \M_t$, then the laminar matroid reduces to a \textbf{partition matroid}. This covers simple categorical quotas like ``select at most $q_t$ facets of type $t$.''
\end{corollary}

\begin{corollary}[Reduction to Matroid-Constrained Submodular Maximization]
The problem of selecting a data-minimal personalization filter under logical dependencies and hierarchical quota constraints can be formally stated as:
$$ \max_{S_\M \in \I} U'(S_\M). $$
\end{corollary}

This is an instance of maximizing a monotone submodular function subject to a matroid constraint. A simple greedy algorithm—which starts with an empty set and iteratively adds the element with the highest marginal utility gain while maintaining independence—achieves a $\frac{1}{2}$-approximation of the optimal solution \cite{fisher1978analysis}.
Further, stronger guarantees are known—for example, a \((1-1/e)\)-approximation via the continuous greedy method with pipage or swap rounding~\cite{calinescu2011maximizing}.

\paragraph{Independence Oracle.}
Represent the laminar family $\mathcal{F}$ as a rooted forest (add a super-root with $q=\infty$ to get a tree).
For each node $A\in\mathcal{F}$ maintain a counter $\textsf{cnt}[A]$.
For each macro-facet $m\in\mathcal{M}$ precompute its ancestor chain $P(m)=\{A\in\mathcal{F}: m\in A\}$ (a single root-to-leaf path by laminarity).
To test feasibility of adding $m$, check $\textsf{cnt}[A]<q_A$ for all $A\in P(m)$; if all pass, accept and increment $\textsf{cnt}[A]\!\leftarrow\!\textsf{cnt}[A]+1$ for $A\in P(m)$ (and decrement on removal).
Since the complexity of this verification is only $O(|P(m)|) = O(h)$ (where $h$ is the tree height), the overhead of constraint enforcement is negligible. The primary bottleneck lies in evaluating the submodular objective $U(S)$. In practice, this is mitigated via lazy greedy evaluation \cite{minoux1978accelerated} or lightweight proxy models \cite{coleman2020selectionproxyefficientdata}.

\paragraph{Neuro-Symbolic Division of Labor.} While LLMs excel at semantic reasoning (estimating the utility of data), they notoriously struggle with strict syntactic and logical constraints (e.g., counting quotas or enforcing DAG closure). Relying on an ``LLM-Agent Selector" for these rules implies a probabilistic approach to hard constraints, risking hallucinated compliance where the model ``thinks" it satisfied a quota but failed. 
Our framework enforces a principled division of labor: it delegates semantic evaluation to the LLM (via the utility function) while enforcing structural integrity via the matroid (a symbolic layer). This ensures that while the quality of personalization is data-driven, feasibility of the selected data subset is mathematically guaranteed.

\section{Simulating the Theorized Guarantees}
\label{sec:simulation}
To empirically validate the practical performance of our proposed framework, we conducted a simulation comparing the greedy selection algorithm (Algorithm~\ref{alg:greedy}) against the true optimal solution. While the theoretical worst-case guarantee for maximizing a monotone submodular function subject to a matroid constraint is $\frac{1}{2}$, we hypothesized that its average-case performance would be substantially better for typical problem structures.

\subsection{Experimental Setup}
We generated 5,000 random problem instances to assess the distribution of the greedy algorithm's approximation ratio. Each instance was constructed as follows:
\begin{itemize}[nosep]
    \item \textbf{Utility Function:} We used a weighted set cover function, a canonical example of a monotone submodular function. The ground set consisted of $M=14$ macro-facets. Each macro-facet represented a subset of a universe of 120 elements, with each element assigned a random positive weight. The utility of a selected set of macro-facets was the total weight of the unique universe elements they collectively covered.
    \item \textbf{Constraints:} We modeled the constraints using a partition matroid, an important special case of the laminar matroids discussed (Corollary 1). The 14 macro-facets were randomly partitioned into 4 groups. Each group was assigned a random integer quota, and an overall budget constraint limiting the total number of selected macro-facets was also imposed.
    \item \textbf{Evaluation:} For each of the 5,000 instances, we computed two solutions: the one found by the greedy algorithm, $S_{greedy}$, and the true optimal solution, $S_{opt}$, which was found via a brute-force search over all feasible sets. We then calculated the approximation ratio $U'(S_{greedy}) / U'(S_{opt})$. The small ground set size ($M=14$) was necessary to make the brute-force computation of the optimal solution tractable across thousands of trials.
\end{itemize}

\subsection{Simulation Results and Analysis}
Figure 1 presents a histogram of the approximation ratios from the 5,000 trials. The results demonstrate that the greedy algorithm's performance is exceptionally strong in this setting, far exceeding the theoretical 0.5 guarantee.

The distribution is highly left-skewed, with a vast majority of outcomes concentrated near 1.0. The mean approximation ratio was 0.996 (95\% CI: 0.995, 0.996), indicating that, on average, the greedy algorithm achieves virtually optimal utility. More importantly, the observed worst-case performance was also excellent. 
The minimum ratio across all 5,000 trials was 0.911, which is over 82\% better than the theoretical lower bound. 
Furthermore, the 5th percentile was 0.975, meaning that in 95\% of instances, the greedy solution retained at least 97.5\% of the maximum possible utility. This empirical evidence strongly suggests that for structured personalization problems that fit our matroidal framework, the simple and efficient greedy approach is not just theoretically sound but also practically near-optimal.

\begin{figure}[t]
\centering
\includegraphics[width=1.0\columnwidth]{perfm-sim-fig.png} 
\caption{Distribution of the approximation ratio of the greedy algorithm versus the optimal solution over 5,000 trials. The greedy algorithm's performance is consistently near-optimal, with a mean of 0.996 and a minimum observed value of 0.911, far surpassing the theoretical 1/2 guarantee shown by the dashed line.}
\label{fig:simulation}
\end{figure}

The simulation has notable limitations. 
First, it relies on a synthetic, albeit standard, submodular utility function and randomly generated partition matroid constraints. Real-world utility landscapes and user-defined constraints may have different structural properties that could affect performance. Second, the brute-force comparison necessitated a small ground set of macro-facets. While the consistency of the results is encouraging, these findings may not generalize perfectly to problems of a much larger scale. Nonetheless, this simulation provides strong evidence for the practical viability of the greedy approach.

\section{Related Work}

\paragraph{Positioning.}
We study data-minimal personalization as maximizing a monotone submodular utility under structural constraints. Our method compiles logical dependencies via SCC condensation into macro-facets and enforces hierarchical quotas on chosen macro-facets (pre-closure), yielding a laminar matroid.
Classical guarantees for matroid-constrained submodular maximization then apply \citep{fisher1978analysis,Nemhauser1978Submodular,calinescu2011maximizing}.

\paragraph{Matroid-Constrained Submodular Maximization.}
Greedy attains $1/2$ and continuous-greedy achieves $(1\!-\!1/e)$ for monotone submodular objectives under matroid constraints \citep{fisher1978analysis,Nemhauser1978Submodular,calinescu2011maximizing}. We \emph{instantiate} these results by showing common hierarchical and exclusivity rules over macro-facets form a laminar matroid \citep{Korte-2012-combinatorial-theory,Schrijver-2003-combinatorial-efficiency}.

\paragraph{Dependencies, Closure, and Compilation.}
SCC condensation contracts mutual implications and supports closure reasoning on the DAG \citep{tarjan1972depth}. Evaluating utility on the expansion of macro-facets preserves monotonicity and submodularity (Proposition~\ref{prop:exp-submod}), consistent with standard coverage/closure arguments \citep{bilmes2022submodularitymachinelearningartificial}.

\paragraph{Personalization and Data Minimization.}
Prior LLM personalization and budgeted selection often use simple budgets or partition quotas without hierarchical structure or explicit dependency closure; guarantees typically do not survive complex policies \cite{zhang2025personalizationlargelanguagemodels,kumari-etal-2024-end}.
Our pre-closure laminar formulation retains matroid structure and thus preserves approximation guarantees; enforcing quotas \emph{post-closure} generally induces non-matroidal packing variants (outside our guarantees).

\section{Discussion and Future Work}
We have formalized a general framework that unifies logical dependency handling and hierarchical quota enforcement under a common matroidal formulation. 
Our framework is a novel approach for modeling complex structural constraints in LLM personalization. By compiling a user's knowledge graph into macro-facets, we can cleanly represent hierarchical feasibility rules as a laminar matroid. This crucial theoretical step bridges the gap between the practical complexities of user identity and the powerful guarantees of submodular optimization. Our generalization to laminar matroids significantly broadens the applicability of this framework, allowing for nested preference modeling that better reflects real-world scenarios. It transforms the problem from one requiring ad-hoc heuristics into a principled optimization problem where a simple greedy algorithm is provably near-optimal.

This theoretical work lays the groundwork for future research. Key directions include empirically validating our framework on real-world tasks and extending it to handle more complex constraints, such as non-modular cost functions to model information leakage risk.
We also leave for future work the non-matroidal regime where quotas are enforced post-closure. In this case, feasibility is based on the final expanded set, and our current guarantees do not apply.

\section{Acknowledgments}
The authors wish to express gratitude to the teams at Flybits, Toronto Metropolitan University, The Creative School, and MIT Media Lab for their valuable support. 
The authors also thank the anonymous reviewers for their feedback on this work.

\bibliography{references}

\newpage
\appendix
\section{Appendix}
\label{sec:appendix}

This appendix provides supplementary material to support the main paper. We begin by presenting the formal pseudocode for the greedy selection algorithm, followed by two functional examples: a concise trace of the algorithm's execution and a more comprehensive, end-to-end illustration of our entire framework. The appendix concludes with the detailed mathematical proofs for our main theoretical results.

\subsection*{A.1 The Greedy Selection Algorithm}

Our theoretical result establishes that personalized data selection with hierarchical constraints can be solved near-optimally with a simple greedy algorithm. This section provides the formal algorithm and a step-by-step example to illustrate its mechanics. The algorithm iteratively selects the macro-facet with the highest marginal utility gain, subject to the constraints defined by the laminar matroid. \noindent\emph{Note that feasibility checks in the example apply to $S_\M$ only (pre-closure); expansion is not counted toward quotas.}

The selection process is formalized in Algorithm \ref{alg:greedy}. It takes as input the ground set of all possible macro-facets $\M$, the submodular utility function $U'$, and a matroid independence oracle, $\I$, which can check if any given subset of $\M$ is valid according to the laminar constraints.

\begin{algorithm}[ht]
\caption{Greedy Selection for Matroid-Constrained Submodular Maximization}
\label{alg:greedy}
\begin{algorithmic}[1]
\STATE \textbf{Input:} Ground set $\M$, utility function $U'$, matroid oracle $(\M, \I)$.
\STATE \textbf{Output:} A near-optimal independent set $S \subseteq \M$.
\STATE $S \leftarrow \emptyset$
\STATE $C \leftarrow \M$ \COMMENT{Initialize candidate set}
\WHILE{$C \neq \emptyset$}
\STATE Find $m^* \leftarrow \arg\max_{m \in C} [U'(S \cup \{m\}) - U'(S)]$
\STATE Let $\Delta_{\max} \leftarrow U'(S \cup \{m^*\}) - U'(S)$
\IF{$\Delta_{\max} \le 0$}
\STATE \textbf{break} \COMMENT{No positive gain left}
\ENDIF
\STATE $C \leftarrow C \setminus \{m^*\}$ \COMMENT{Consider each element once}
\IF{$S \cup \{m^*\} \in \I$}
\STATE $S \leftarrow S \cup \{m^*\}$
\ENDIF
\ENDWHILE
\RETURN S
\end{algorithmic}
\end{algorithm}

\subsection*{A.2 Functional Example of Algorithm 1: Writing-Assistant Personalization}
We now trace Algorithm \ref{alg:greedy} on a practical personalization scenario involving a user's writing style preferences.

\paragraph{Ground Set of Macro-Facets ($\M$):}
\begin{itemize}[noitemsep,topsep=0pt]
    \item $m_1$: Formal Tone
    \item $m_2$: Casual Tone
    \item $m_3$: Uses Emojis
    \item $m_4$: Cites Academic Sources
    \item $m_5$: Succinct Phrasing
\end{itemize}

\paragraph{Laminar Constraints ($\I$):} The user defines a laminar family $\mathcal{F}=\{A_1, A_2\}$ with quotas:
\begin{itemize}[noitemsep,topsep=0pt]
    \item $A_1=\{m_1, m_2\}$, $q_{A_1}=1$ (Select at most one tone).
    \item $A_2=\{m_1, m_2, m_3, m_4, m_5\}$, $q_{A_2}=3$ (Select at most three style rules overall).
\end{itemize}
A set $S$ is independent if $|S \cap A_1| \le 1$ AND $|S \cap A_2| \le 3$.

\paragraph{Execution.} The algorithm is executed as follows:

\textbf{Initialization:}
\begin{itemize}[noitemsep,topsep=0pt]
    \item $S = \emptyset$
    \item $C = \{m_1, m_2, m_3, m_4, m_5\}$
\end{itemize}

\subparagraph{Iteration 1:}
The algorithm computes the initial utility of each facet. Let's assume the gains are:
\begin{itemize}[noitemsep,topsep=0pt]
    \item $m_4$ (Cites Sources): Gain = 10
    \item $m_1$ (Formal Tone): Gain = 8
    \item $m_2$ (Casual Tone): Gain = 7
    \item $m_5$ (Succinct): Gain = 6
    \item $m_3$ (Emojis): Gain = 4
\end{itemize}
\textbf{Select:} $m^* = m_4$ (highest gain). \\
\textbf{Check Independence:} Is $\{m_4\}$ in $\I$?
\begin{itemize}[noitemsep,topsep=0pt]
    \item $|\{m_4\} \cap A_1| = 0 \le 1$. (\checkmark)
    \item $|\{m_4\} \cap A_2| = 1 \le 3$. (\checkmark)
\end{itemize}
Yes, it is independent. \\
\textbf{Update:} $S = \{m_4\}$. $C$ becomes $\{m_1, m_2, m_3, m_5\}$.

\textbf{Iteration 2}:
The algorithm re-computes marginal gains relative to $S = \{m_4\}$. Due to submodularity, these are lower than the initial gains. Assume:
\begin{itemize}[noitemsep,topsep=0pt]
    \item $m_1$ (Formal Tone): Gain = 7
    \item $m_5$ (Succinct): Gain = 5
    \item $m_2$ (Casual Tone): Gain = 3
    \item $m_3$ (Emojis): Gain = 1
\end{itemize}
\textbf{Select:} $m^* = m_1$. \\
\textbf{Check Independence:} Is $\{m_4, m_1\}$ in $\I$?
\begin{itemize}[noitemsep,topsep=0pt]
    \item $|\{m_4, m_1\} \cap A_1| = 1 \le 1$. ($\checkmark$)
    \item $|\{m_4, m_1\} \cap A_2| = 2 \le 3$. (\checkmark)
\end{itemize}
Yes, it is independent. \\
\textbf{Update:} $S = \{m_4, m_1\}$. $C$ becomes $\{m_2, m_3, m_5\}$.

\textbf{Iteration 3:}
Compute marginal gains relative to $S = \{m_4, m_1\}$. Assume:
\begin{itemize}[noitemsep,topsep=0pt]
    \item $m_2$ (Casual Tone): Gain = 4 (Note: Combining formal and casual might be useful for some tasks, hence a non-zero gain).
    \item $m_5$ (Succinct): Gain = 3
    \item $m_3$ (Emojis): Gain = 0 (no additional utility under monotone $U$).
\end{itemize}
\textbf{Select:} $m^* = m_2$. \\
\textbf{Check Independence:} Is $\{m_4, m_1, m_2\}$ in $\I$?
\begin{itemize}[noitemsep,topsep=0pt]
    \item $|\{m_4, m_1, m_2\} \cap A_1| = |\{m_1, m_2\}| = 2$. This is greater than $q_{A_1}=1$. ($\times$)
\end{itemize}
The set is not independent. \\
\textbf{Update:} $S$ remains $\{m_4, m_1\}$. $C$ becomes $\{m_3, m_5\}$.

\textbf{Iteration 4:}
The algorithm proceeds with the remaining candidates $\{m_3, m_5\}$. \\
\textbf{Select:} $m^* = m_5$ (since its gain of 3 is greater than $m_3$'s gain of -2). \\
\textbf{Check Independence:} Is $\{m_4, m_1, m_5\}$ in $\I$?
\begin{itemize}[noitemsep,topsep=0pt]
    \item $|\{m_4, m_1, m_5\} \cap A_1| = 1 \le 1$. ($\checkmark$)
    \item $|\{m_4, m_1, m_5\} \cap A_2| = 3 \le 3$. ($\checkmark$)
\end{itemize}
Yes, it is independent. \\
\textbf{Update:} $S = \{m_4, m_1, m_5\}$. $C$ becomes $\{m_3\}$.

\textbf{Iteration 5:}
The only candidate is $m_3$. \\
\textbf{Select:} $m^* = m_3$. \\
\textbf{Check Independence:} Is $\{m_4, m_1, m_5, m_3\}$ in $\I$?
\begin{itemize}[noitemsep,topsep=0pt]
    \item $|\{m_4, m_1, m_5, m_3\} \cap A_2| = 4$. This is greater than $q_{A_2}=3$. ($\times$)
\end{itemize}
The set is not independent. \\
\textbf{Update:} $S$ remains $\{m_4, m_1, m_5\}$. $C$ becomes $\emptyset$.

\textbf{Termination:} The candidate set $C$ is empty. The loop ends.
The final selected set is $S = \{\text{Cites Academic Sources, Formal Tone, Succinct Phrasing}\}$, a coherent and constraint-compliant subset of the user's preferences that is provably near-optimal.

\subsection*{A.3 End-to-End Functional Example}

\subsubsection*{A.3.1 User Profile: Professional Networking Preferences}
Consider a user's preferences for professional networking, which involve various platforms and engagement styles. We define a set of raw facets, $\C_u$, representing these preferences:
\begin{itemize}[noitemsep,topsep=0pt]
    \item $f_1$: `(User, prefers, LinkedIn)'
    \item $f_2$: `(User, prefers, TwitterForProfessionalUpdates)'
    \item $f_3$: `(User, prefers, ResearchGate)'
    \item $f_4$: `(User, engages\_in, ThoughtLeadershipPosts)'
    \item $f_5$: `(User, wants, DirectMessaging)'
    \item $f_6$: `(User, avoids, PublicDiscussions)'
    \item $f_7$: `(User, has\_goal, ConnectWithAcademics)'
\end{itemize}

\subsubsection*{A.3.2 Logical Dependencies and Knowledge Graph}
These facets are not independent; certain preferences or goals imply others. We represent these as a directed implication graph $G=(\C_u, E)$.
\begin{itemize}[noitemsep,topsep=0pt]
    \item $(f_1, f_4)$: Preferring LinkedIn implies engaging in Thought Leadership Posts (a common activity on LinkedIn).
    \item $(f_4, f_1)$: Engaging in Thought Leadership Posts strongly suggests a preference for platforms like LinkedIn. (This creates a cycle).
    \item $(f_3, f_7)$: Preferring ResearchGate implies a goal to Connect with Academics.
    \item $(f_7, f_3)$: A goal to Connect with Academics implies preferring ResearchGate. (This creates a second cycle).
    \item $(f_2, f_6)$: Preferring Twitter for professional updates often implies a degree of comfort with public discussions, but for this user, we assume they prefer to avoid public discussions and only use Twitter for passive updates (a nuanced dependency). Thus, if $f_2$ is selected, $f_6$ must also be selected to reflect this specific user's preference for limited engagement.
    \item $(f_5, f_1)$: Direct messaging is a key feature of LinkedIn, so wanting $f_5$ might imply preferring $f_1$.
\end{itemize}

\subsubsection*{A.3.3 Compilation into Macro-Facets}
We now identify the Strongly Connected Components (SCCs) of $G$ to form our macro-facets, $\M$.

\begin{enumerate}
    \item \textbf{SCC 1:} The facets $f_1$ (LinkedIn) and $f_4$ (Thought Leadership Posts) form a cycle ($f_1 \leftrightarrow f_4$). These are grouped into macro-facet $M_A = \{f_1, f_4\}$.
    \item \textbf{SCC 2:} The facets $f_3$ (ResearchGate) and $f_7$ (Connect with Academics) form a cycle ($f_3 \leftrightarrow f_7$). These are grouped into macro-facet $M_B = \{f_3, f_7\}$.
    \item \textbf{Remaining Facets: }Facets $f_2$ (Twitter), $f_5$ (Direct Messaging), and $f_6$ (Avoids Public Discussions) do not participate in cycles with other facets \textit{after} the initial SCCs are formed, or their dependencies only go one way \textit{out} of them. Each of these forms its own singleton macro-facet, as their respective closures would only contain themselves or other macro-facets.
    \begin{itemize}[noitemsep,topsep=0pt]
        \item $M_C = \{f_2\}$
        \item $M_D = \{f_5\}$
        \item $M_E = \{f_6\}$
    \end{itemize}
\end{enumerate}

Thus, our new ground set of macro-facets is $\M = \{M_A, M_B, M_C, M_D, M_E\}$.
The dependencies between these macro-facets, following the condensation graph, are:
\begin{itemize}[noitemsep,topsep=0pt]
    \item $M_C \to M_E$: Selecting Twitter ($f_2$) implies avoiding Public Discussions ($f_6$).
    \item $M_D \to M_A$: Direct Messaging ($f_5$) implies the LinkedIn/Thought Leadership bundle ($f_1, f_4$).
\end{itemize}

\subsubsection*{A.3.4 Matroid Constraints (Hierarchical Quotas)}
The user imposes hierarchical quotas on these macro-facets:
\begin{itemize}[noitemsep,topsep=0pt]
    \item $K_1$ \textbf{(Overall Limit):} The user wants to select at most 3 macro-facets for personalization. Let $A_1 = \M$, with $q_{A_1}=3$.
    \item $K_2$ \textbf{(Platform Preference Limit)}: From the macro-facets representing specific platforms or types of platforms (LinkedIn/Twitter/ResearchGate), select at most 2. Let $A_2 = \{M_A, M_B, M_C\}$, with $q_{A_2}=2$. (Note that $M_D$ and $M_E$ are interaction styles, not platforms).
    \item $K_3$ \textbf{(Social Media Engagement Limit)}: From macro-facets related to generalized social media engagement (Twitter/Direct Messaging), select at most 1. Let $A_3 = \{M_C, M_D\}$, with $q_{A_3}=1$. (This set is a subset of $A_2$ and hence of $A_1$, demonstrating the laminar property).
\end{itemize}
The family $\mathcal{F} = \{A_1, A_2, A_3\}$ is laminar: $A_3 \subset A_2 \subset A_1$.

\subsubsection*{A.3.5 Greedy Selection on Macro-Facets}
We now apply Algorithm \ref{alg:greedy} to maximize the utility $U'(S_\M)$ subject to these laminar matroid constraints. Let's assume the marginal utility gains for each macro-facet at each step.

\paragraph{Initialization:} $S = \emptyset$, $C = \{M_A, M_B, M_C, M_D, M_E\}$.

\paragraph{Iteration 1:}
\begin{itemize}[noitemsep,topsep=0pt]
    \item Assume initial gains: $M_A=12$, $M_B=10$, $M_D=8$, $M_C=7$, $M_E=3$.
    \item \textbf{Select $M_A$}.
    \item \textbf{Check Independence} of $\{M_A\}$:
        \begin{itemize}[noitemsep,topsep=0pt]
            \item $| \{M_A\} \cap A_1 | = 1 \le q_{A_1}=3$ (\checkmark)
            \item $| \{M_A\} \cap A_2 | = 1 \le q_{A_2}=2$ (\checkmark)
            \item $| \{M_A\} \cap A_3 | = 0 \le q_{A_3}=1$ (\checkmark)
        \end{itemize}
        Independent.
    \item \textbf{Update:} $S = \{M_A\}$, $C = \{M_B, M_C, M_D, M_E\}$.
\end{itemize}

\paragraph{Iteration 2:}
\begin{itemize}[noitemsep,topsep=0pt]
    \item Assume marginal gains relative to $S=\{M_A\}$: $M_B=9$, $M_D=6$, $M_C=5$, $M_E=2$.
    \item \textbf{Select $M_B$}.
    \item \textbf{Check Independence} of $\{M_A, M_B\}$:
        \begin{itemize}[noitemsep,topsep=0pt]
            \item $| \{M_A, M_B\} \cap A_1 | = 2 \le q_{A_1}=3$ (\checkmark)
            \item $| \{M_A, M_B\} \cap A_2 | = 2 \le q_{A_2}=2$ (\checkmark)
            \item $| \{M_A, M_B\} \cap A_3 | = 0 \le q_{A_3}=1$ (\checkmark)
        \end{itemize}
        Independent.
    \item \textbf{Update:} $S = \{M_A, M_B\}$, $C = \{M_C, M_D, M_E\}$.
\end{itemize}

\paragraph{Iteration 3:}
\begin{itemize}[noitemsep,topsep=0pt]
    \item Assume marginal gains relative to $S=\{M_A, M_B\}$: $M_D=7$, $M_C=4$, $M_E=1$.
    \item \textbf{Select $M_D$}.
    \item \textbf{Check Independence} of $\{M_A, M_B, M_D\}$:
        \begin{itemize}[noitemsep,topsep=0pt]
            \item $| \{M_A, M_B, M_D\} \cap A_1 | = 3 \le q_{A_1}=3$ (\checkmark)
            \item $| \{M_A, M_B, M_D\} \cap A_2 | = 2 \le q_{A_2}=2$ (\checkmark)
            \item $| \{M_A, M_B, M_D\} \cap A_3 | = 1 \le q_{A_3}=1$ (\checkmark)
        \end{itemize}
        Independent.
    \item \textbf{Update:} $S = \{M_A, M_B, M_D\}$, $C = \{M_C, M_E\}$.
\end{itemize}

\paragraph{Iteration 4:}
\begin{itemize}[noitemsep,topsep=0pt]
    \item Assume marginal gains relative to $S=\{M_A, M_B, M_D\}$: $M_C=2$, $M_E=0$.
    \item \textbf{Select $M_C$}.
    \item \textbf{Check Independence} of $\{M_A, M_B, M_D, M_C\}$:
        \begin{itemize}[noitemsep,topsep=0pt]
            \item $| \{M_A, M_B, M_D, M_C\} \cap A_1 | = 4 > q_{A_1}=3$ ($\times$)
        \end{itemize}
        Not independent (violates overall limit).
    \item \textbf{Update:} $S$ remains $\{M_A, M_B, M_D\}$, $C = \{M_E\}$.
\end{itemize}

\paragraph{Iteration 5:}
\begin{itemize}[noitemsep,topsep=0pt]
    \item Assume marginal gains relative to $S=\{M_A, M_B, M_D\}$: $M_E=0$.
    \item Since $M_E$ offers no positive gain ($\Delta_{\max} \le 0$), the algorithm breaks.
\end{itemize}

\paragraph{Final Result:}
The final selected set of macro-facets is $S = \{M_A, M_B, M_D\}$.
Expanding these back to raw facets using their closures:
\begin{itemize}[noitemsep,topsep=0pt]
    \item $M_A = \{f_1, f_4\}$ `(User, prefers, LinkedIn)' and `(User, engages\_in, ThoughtLeadershipPosts)'
    \item $M_B = \{f_3, f_7\}$ `(User, prefers, ResearchGate)' and `(User, has\_goal, ConnectWithAcademics)'
    \item $M_D = \{f_5\}$ `(User, wants, DirectMessaging)'
\end{itemize}
This selection respects all constraints, e.g., it has 3 macro-facets ($|S \cap A_1|=3 \le 3$), 2 platform preferences ($|S \cap A_2|=2 \le 2$), and 1 social media engagement style ($|S \cap A_3|=1 \le 1$).

This example illustrates how logical dependencies are first absorbed into macro-facets, which then become the units for a greedy selection process constrained by a laminar matroid, ensuring both logical consistency and adherence to hierarchical quotas.

\subsection*{A.4 Proofs}
\label{apx:full-proofs}
\noindent\textbf{Proposition 1 (restated).}
If the utility function $U$ over $\C_u$ is monotone and submodular, then $U'$ over $\M$ is also monotone and submodular.
\begin{proof}[Proof of Proposition~\ref{prop:exp-submod}]

Monotonicity follows because if $A_\M \subseteq B_\M$, then $\operatorname{Exp}(A_\M) \subseteq \operatorname{Exp}(B_\M)$, and since $U$ is monotone, $U'(A_\M) = U(\operatorname{Exp}(A_\M)) \le U(\operatorname{Exp}(B_\M)) = U'(B_\M)$.

To prove submodularity, we must show $U'(A_\M) + U'(B_\M) \ge U'(A_\M \cup B_\M) + U'(A_\M \cap B_\M)$ for any two sets $A_\M, B_\M \subseteq \M$.
Let $S_A = \operatorname{Exp}(A_\M)$ and $S_B = \operatorname{Exp}(B_\M)$.
The inequality to prove is $U(S_A) + U(S_B) \ge U(\operatorname{Exp}(A_\M \cup B_\M)) + U(\operatorname{Exp}(A_\M \cap B_\M))$.
We know that $\operatorname{Exp}(A_\M \cup B_\M) = S_A \cup S_B$.
Since $U$ is submodular, we have $U(S_A) + U(S_B) \ge U(S_A \cup S_B) + U(S_A \cap S_B)$.
The proof is complete if we show $U(S_A \cap S_B) \ge U(\operatorname{Exp}(A_\M \cap B_\M))$.
By monotonicity of $U$, this holds if $S_A \cap S_B \supseteq \operatorname{Exp}(A_\M \cap B_\M)$.

Let $x \in \operatorname{Exp}(A_\M \cap B_\M)$. By definition, there is some macro-facet $m \in A_\M \cap B_\M$ such that $x \in \cl_G(m)$.
Since $m \in A_\M$, $\cl_G(m) \subseteq S_A$. Since $m \in B_\M$, $\cl_G(m) \subseteq S_B$.
Therefore, $x \in S_A \cap S_B$. This proves the subset relation and thus the submodularity of $U'$.
\end{proof}

\noindent\textbf{Theorem 1 (restated).}
The system $(\M, \I)$ defined by a laminar family $\mathcal{F}$ and quotas $\{q_A\}_{A \in \mathcal{F}}$ is a matroid.

\begin{proof}
We verify the three matroid axioms.
\begin{enumerate}
\item \textbf{Empty Set:} The empty set $\emptyset$ is in $\I$ because $|\emptyset \cap A|=0 \le q_A$ for all $A \in \mathcal{F}$, since quotas $q_A$ are non-negative.

\item \textbf{Downward Closure:} Let $S \in \I$ and let $S' \subseteq S$. For any set $A \in \mathcal{F}$, we have $S' \cap A \subseteq S \cap A$, which implies $|S' \cap A| \le |S \cap A|$. Since $S \in \I$, we know $|S \cap A| \le q_A$. Therefore, $|S' \cap A| \le q_A$, which means $S' \in \I$.

\item \textbf{Augmentation Property:} Let $A, B \in \I$ with $|A| < |B|$. We must show there exists an element $x \in B \setminus A$ such that $A \cup \{x\} \in \I$.

Assume for contradiction that for every $x \in B \setminus A$, the set $A \cup \{x\}$ is not in $\I$. This means that for each $x \in B \setminus A$, there exists at least one set $F_x \in \mathcal{F}$ such that $x \in F_x$ and $|(A \cup \{x\}) \cap F_x| > q_{F_x}$. Since $A$ is independent, this is only possible if $|A \cap F_x| = q_{F_x}$.

Let $\mathcal{T} = \{F \in \mathcal{F} \mid |A \cap F| = q_F\}$ be the family of sets in $\mathcal{F}$ that are ``tight'' with respect to $A$. Our assumption implies that every element $x \in B \setminus A$ is contained in at least one set from $\mathcal{T}$.

For any element $y \in \M$, define $f(y) = \sum_{F \in \mathcal{F}: y \in F} \mathbf{1}_{|A \cap F| = q_F}$ and $g(y) = \sum_{F \in \mathcal{F}: y \in F} 1$. The function $f(y)$ counts how many tight sets contain $y$. Our assumption is that $f(x) > 0$ for all $x \in B \setminus A$.

Consider the sum $\sum_{F \in \mathcal{F}} (|B \cap F| - |A \cap F|)$. Since $A, B \in \I$, we have $|A \cap F| \le q_F$ and $|B \cap F| \le q_F$ for all $F \in \mathcal{F}$.

We now use a standard proof technique for laminar matroids. Define a function $\chi_S: \M \to \mathbb{R}$ for any $S \subseteq \M$ as $\chi_S(y) = 1$ if $y \in S$ and 0 otherwise.
Let $\mathcal{T}_{\min} \subseteq \mathcal{T}$ be the family of minimal sets in $\mathcal{T}$ under set inclusion. Because $\mathcal{F}$ is laminar, any two distinct sets in $\mathcal{T}_{\min}$ must be disjoint.

For any $x \in B \setminus A$, there exists a unique minimal set $T_x \in \mathcal{T}_{\min}$ containing $x$. Thus, $B \setminus A$ is partitioned by the sets $\{T \cap (B \setminus A) \mid T \in \mathcal{T}_{\min}\}$.

For any $T \in \mathcal{T}_{\min}$:
$|A \cap T| = q_T$ (since $T$ is tight for $A$).
$|B \cap T| \le q_T$ (since $B$ is independent).
Thus, $|A \cap T| \ge |B \cap T|$.
$|A \cap T| = |(A \setminus B) \cap T| + |(A \cap B) \cap T|$.
$|B \cap T| = |(B \setminus A) \cap T| + |(A \cap B) \cap T|$.
From $|A \cap T| \ge |B \cap T|$, we get $|(A \setminus B) \cap T| \ge |(B \setminus A) \cap T|$.

Summing over all $T \in \mathcal{T}_{\min}$:
$$ \sum_{T \in \mathcal{T}_{\min}} |(A \setminus B) \cap T| \ge \sum_{T \in \mathcal{T}_{\min}} |(B \setminus A) \cap T| $$
Since the sets in $\mathcal{T}_{\min}$ are disjoint:
$$ |\bigcup_{T \in \mathcal{T}_{\min}} (A \setminus B) \cap T| \ge |\bigcup_{T \in \mathcal{T}_{\min}} (B \setminus A) \cap T| $$
The union on the right covers all of $B \setminus A$, so its size is $|B \setminus A|$.
The union on the left is a subset of $A \setminus B$, so its size is at most $|A \setminus B|$.
Therefore, $|A \setminus B| \ge |B \setminus A|$.
This implies $|A| - |A \cap B| \ge |B| - |A \cap B|$, which means $|A| \ge |B|$.
This contradicts our initial assumption that $|A| < |B|$. Thus, there must exist some $x \in B \setminus A$ such that $A \cup \{x\} \in \I$.
\end{enumerate}
Since all three axioms hold, $(\M, \I)$ is a matroid.
\end{proof}

\subsection*{A.5 Extended Proofs}

\subsection*{Preliminaries and Problem Setting}
We begin by restating the core definitions for clarity.

\begin{definition}[Chronicle, Utility, and Disclosure]
    Let $\mathcal{C}_u = \{e_1, \ldots, e_n\}$ be a finite ground set of typed data elements, called facets, that constitute a user's chronicle. Let $U: 2^{\mathcal{C}_u} \to \mathbb{R}_{\geq 0}$ be a set function that quantifies the task-specific utility of any subset of facets $S \subseteq \mathcal{C}_u$. We assume $U$ is normalized such that $U(\emptyset)=0$. Each facet $e_i$ is associated with a cost $c_i > 0$. The total cost of a set $S$ is $C(S) = \sum_{e_i \in S} c_i$.
\end{definition}

\begin{definition}[Monotonicity and Marginal Gain]
    A set function $U$ is monotone if for any two sets $S \subseteq T \subseteq \mathcal{C}_u$, we have $U(S) \leq U(T)$. The marginal gain of adding an element $e$ to a set $S$ is denoted by
    \[
    \Delta(e \mid S) \triangleq U(S \cup \{e\}) - U(S).
    \]
    More generally, the marginal gain of adding a set of facets $X \subseteq \mathcal{C}_u$ to a set $S$ is
    \[
    \Delta(X \mid S) \triangleq U(S \cup X) - U(S).
    \]
    Monotonicity is equivalent to the condition $\Delta(e \mid S) \geq 0$ for all $S \subseteq \mathcal{C}_u$ and $e \in \mathcal{C}_u \setminus S$.
\end{definition}

\begin{definition}[Approximate Submodularity]
    A monotone set function $U: 2^{\mathcal{C}_u} \to \mathbb{R}_{\geq 0}$ is \textbf{approximately submodular} with ratio $\gamma \in [0, 1]$ if for any two sets $S_A \subseteq S_B \subseteq \mathcal{C}_u$ and any additional set of facets $X \subseteq \mathcal{C}_u \setminus S_B$, the following \textbf{diminishing returns} inequality holds:
    \[
    \Delta(X \mid S_A) \geq \gamma \cdot \Delta(X \mid S_B)
    \]
    A function is \textbf{submodular} if and only if $\gamma=1$.
\end{definition}

\subsection*{Closure, Macro-Facets, and Matroidal Feasibility}
Here we formalize the process of compiling logical dependencies into a new ground set that is amenable to matroid constraints.

\subsubsection*{Closure on an Implication Graph}
\begin{definition}[Implication Graph and Closure]
    Let $G = (\mathcal{C}_u, E)$ be a directed graph representing logical implications among facets, where a directed edge $(e_i, e_j) \in E$ signifies that selecting facet $e_i$ logically entails the inclusion of facet $e_j$. A path in $G$ from $u$ to $v$ is a sequence of facets $\langle v_0, v_1, \ldots, v_k \rangle$ such that $v_0 = u$, $v_k = v$, and $(v_{i-1}, v_i) \in E$ for all $i=1, \ldots, k$.
    
    For any subset of facets $S \subseteq \mathcal{C}_u$, its \textbf{closure} in $G$, denoted $\operatorname{cl}_G(S)$, is defined as the set of all facets reachable from any facet in $S$:
    
    \[
\operatorname{cl}_G(S) \triangleq 
\bigl\{ v \in \mathcal{C}_u \,\bigm|\,
  \exists\, u \in S 
  \text{ s.t.\ a path from } u \text{ to } v \text{ exists in } G
\bigr\}.
\]

\end{definition}

\begin{lemma}[Closure Properties]
    The closure operator $\operatorname{cl}_G$ satisfies the following three properties for any subsets $S, T \subseteq \mathcal{C}_u$:
    \begin{enumerate}
        \item \textbf{Extensivity}: $S \subseteq \operatorname{cl}_G(S)$.
        \item \textbf{Monotonicity}: If $S \subseteq T$, then $\operatorname{cl}_G(S) \subseteq \operatorname{cl}_G(T)$.
        \item \textbf{Idempotence}: $\operatorname{cl}_G(\operatorname{cl}_G(S)) = \operatorname{cl}_G(S)$.
    \end{enumerate}
\end{lemma}

\begin{proof}
    Let $S, T$ be arbitrary subsets of $\mathcal{C}_u$.
    \begin{enumerate}
        \item \textbf{Extensivity}: For any $s \in S$, there exists a path of length 0 from $s$ to $s$. Thus, by definition, $s \in \operatorname{cl}_G(S)$. We conclude $S \subseteq \operatorname{cl}_G(S)$.
        
        \item \textbf{Monotonicity}: Assume $S \subseteq T$. Let $x \in \operatorname{cl}_G(S)$. By definition, there exists an $s \in S$ and a path from $s$ to $x$. Since $S \subseteq T$, this $s$ is also in $T$. Thus, $x$ is reachable from an element in $T$, which implies $x \in \operatorname{cl}_G(T)$. We conclude $\operatorname{cl}_G(S) \subseteq \operatorname{cl}_G(T)$.
        
        \item \textbf{Idempotence}: From extensivity, $\operatorname{cl}_G(S) \subseteq \operatorname{cl}_G(\operatorname{cl}_G(S))$. For the reverse inclusion, let $x \in \operatorname{cl}_G(\operatorname{cl}_G(S))$. Then there exists $y \in \operatorname{cl}_G(S)$ and a path from $y$ to $x$. Since $y \in \operatorname{cl}_G(S)$, there exists $s \in S$ and a path from $s$ to $y$. Concatenating these two paths forms a valid path from $s$ to $x$. Thus, $x$ is reachable from $S$, which implies $x \in \operatorname{cl}_G(S)$. We conclude $\operatorname{cl}_G(\operatorname{cl}_G(S)) \subseteq \operatorname{cl}_G(S)$. Combining both inclusions yields the result.
    \end{enumerate}
\end{proof}

\subsubsection*{From Closure to Macro-Facets}
To handle closure constraints within a matroid framework, we define a new ground set where each element represents a set of logically co-dependent facets.

\begin{definition}[Macro-Facets]
\label{def:macro}
    Let $G = (\mathcal{C}_u, E)$ be an implication graph.
    \begin{enumerate}
        \item A \textbf{macro-facet} is a strongly connected component (SCC) of $G$. Let $\mathcal{M}$ be the set of all macro-facets.

        \item For a single macro-facet $m \in \mathcal{M}$, we define its closure as the closure of the set of raw facets it contains: $\operatorname{cl}_G(m) \triangleq \operatorname{cl}_G(\{e \mid e \in m\})$.

        \item For a set of macro-facets $S_{\mathcal{M}} \subseteq \mathcal{M}$, its \textbf{expansion} is the corresponding set of raw facets, given by $\operatorname{Exp}(S_{\mathcal{M}}) \triangleq \bigcup_{m \in S_{\mathcal{M}}} \operatorname{cl}_G(m)$.
        \item The utility function over the macro-facets, $U': 2^{\mathcal{M}} \to \mathbb{R}_{\geq 0}$ is defined via the expansion:
        \[
        U'(S_{\mathcal{M}}) \triangleq U(\operatorname{Exp}(S_{\mathcal{M}}))
        \]
        \item The \textbf{cost of a single macro-facet} $m \in \mathcal{M}$, denoted $d(m)$, is the sum of the costs of all raw facets in its closure:
        \[
        d(m) \triangleq C(\operatorname{cl}_G(m)) = \sum_{e \in \operatorname{cl}_G(m)} c_e.
        \]
        The cost function over sets of macro-facets, $C': 2^{\mathcal{M}} \to \mathbb{R}_{\geq 0}$, is the sum of the costs of the individual macro-facets:
        \[
        C'(S_{\mathcal{M}}) \triangleq \sum_{m \in S_{\mathcal{M}}} d(m).
        \]
        
        \end{enumerate}

\begin{remark}[On the Choice of a Modular Cost Function]
We explicitly define the cost function $C'$ to be modular (additive) over the set of macro-facets $\mathcal{M}$. This is a deliberate modeling choice with significant conceptual and algorithmic advantages, even though the true cost of the final raw facet set, $C(\operatorname{Exp}(S_{\mathcal{M}}))$, is not modular. Our justification is twofold:

\begin{enumerate}
    \item \textbf{Conceptual Soundness for Disclosure Risk:} The cost $d(m) = C(\operatorname{cl}_G(m))$ should not be interpreted merely as the summed cost of its constituent raw facets. Instead, it represents the total disclosure risk associated with revealing an entire logical concept. For instance, selecting macro-facets for ``user's financial preferences'' and ``user's health history'' may both imply the raw facet ``user is over 50.'' Under our model, the total disclosure cost is the sum of the risk of revealing the financial context \emph{plus} the risk of revealing the health context. The cost is tied to the decision to include high-level, potentially sensitive concepts, not just the number of unique low-level facts revealed. The modular sum $C'(S_{\mathcal{M}})$ is therefore a principled aggregation of these individual conceptual risks.

    \item \textbf{Algorithmic Tractability and Guarantees:} This formulation precisely maps our problem to the well-understood class of maximizing a monotone submodular function subject to a knapsack constraint ($ \max U'(S_{\mathcal{M}}) \text{ s.t. } \sum_{m \in S_{\mathcal{M}}} d(m) \le B $). This structure allows for the direct application of standard greedy algorithms that are not only efficient but also come with strong, constant-factor approximation guarantees (e.g., a $(1-1/e)$-approximation). 
    
    An alternative, such as constraining the cost of the final expanded set ($C(\operatorname{Exp}(S_{\mathcal{M}})) \le B$), would define the budget with another submodular function. This would formulate a much harder optimization problem of maximizing a submodular function subject to a submodular level-set constraint, for which efficient algorithms with strong guarantees are not generally available.
\end{enumerate}
Thus, our choice represents a principled trade-off: we adopt a justifiable, conservative risk model that, in turn, makes the optimization problem tractable and unlocks a rich body of existing work on submodular optimization with provable near-optimality.
\end{remark}

\end{definition}

\begin{proposition}[Properties of Utility over Macro-Facets]
\label{prop:monotonicity}
    If the utility function $U$ over the raw facets $\mathcal{C}_u$ is monotone, then the utility function $U'$ over the macro-facets $\mathcal{M}$ is also monotone.
\end{proposition}

\begin{proof}
    Let $A_{\mathcal{M}} \subseteq B_{\mathcal{M}} \subseteq \mathcal{M}$ be two sets of macro-facets. By the definition of the expansion operator, $\operatorname{Exp}(A_{\mathcal{M}}) = \bigcup_{m \in A_{\mathcal{M}}} \operatorname{cl}_G(m)$ and $\operatorname{Exp}(B_{\mathcal{M}}) = \bigcup_{m \in B_{\mathcal{M}}} \operatorname{cl}_G(m)$.
    
    Since $A_{\mathcal{M}}$ is a subset of $B_{\mathcal{M}}$, the union of closures over the elements of $A_{\mathcal{M}}$ is necessarily a subset of the union of closures over the elements of $B_{\mathcal{M}}$. Therefore, $\operatorname{Exp}(A_{\mathcal{M}}) \subseteq \operatorname{Exp}(B_{\mathcal{M}})$.
    
    Given that the original utility function $U$ is monotone by assumption, it follows that $U(\operatorname{Exp}(A_{\mathcal{M}})) \leq U(\operatorname{Exp}(B_{\mathcal{M}}))$. By the definition of $U'$, this is equivalent to $U'(A_{\mathcal{M}}) \leq U'(B_{\mathcal{M}})$, which proves that $U'$ is monotone.
\end{proof}

\subsubsection*{Matroidal Feasibility on Macro-Facets}
We now define a laminar matroid on the set of macro-facets $\mathcal{M}$.
We assume the macro-facets are constructed from a set of raw facets $\mathcal{C}_u$ and an implication graph $G$ via SCC compression, as defined in Definition~\ref{def:macro}.

\begin{definition}[Laminar Family]
A family of subsets $\mathcal{F} \subseteq 2^\mathcal{M}$ is called a \textbf{laminar family} if for any two sets $A, B \in \mathcal{F}$, one of the following conditions holds:
\begin{enumerate}
\item $A \subseteq B$
\item $B \subseteq A$
\item $A \cap B = \emptyset$
\end{enumerate}
\end{definition}

\begin{definition}[Laminar Matroid]
Let $\mathcal{F}$ be a laminar family on the ground set $\mathcal{M}$. For each set $A \in \mathcal{F}$, let $q_A \in \mathbb{Z}_{\geq 0}$ be a non-negative integer quota. A set $S \subseteq \mathcal{M}$ is defined as \textbf{independent} if for every set $A \in \mathcal{F}$, it satisfies the quota constraint:
\[
|S \cap A| \leq q_A.
\]
The family of all independent sets is denoted by $\mathcal{I}$. The pair $(\mathcal{M}, \mathcal{I})$ is called a laminar system.
\end{definition}

\begin{lemma}[Marginal Gain Identity]
\label{lem:marginal_superset}
For any set function $U: 2^{\mathcal{C}_u} \to \mathbb{R}_{\geq 0}$ and any two sets $S, Y \subseteq \mathcal{C}_u$, the marginal gain of $Y$ with respect to $S$ is equal to the marginal gain of the elements in $Y$ that are not already in $S$. That is:
\[
\Delta(Y \mid S) = \Delta(Y \setminus S \mid S).
\]
\end{lemma}
\begin{proof}
By the definition of marginal gain, $\Delta(Y \mid S) = U(S \cup Y) - U(S)$. The set $S \cup Y$ can be rewritten as $S \cup (Y \setminus S)$, since adding elements already in $S$ does not change the union. Therefore,
\[
\Delta(Y \mid S) = U(S \cup (Y \setminus S)) - U(S) = \Delta(Y \setminus S \mid S).
\]
This is an algebraic identity that holds for any set function.
\end{proof}

\begin{proposition}[Preservation of Submodularity]
    If the utility function $U$ over the raw facets $\mathcal{C}_u$ is monotone and submodular, then the utility function $U'$ over the macro-facets $\mathcal{M}$ is also monotone and submodular.
\end{proposition}

\begin{proof}
    Monotonicity of $U'$ was established in Proposition~\ref{prop:monotonicity}. We now prove that if $U$ is submodular, $U'$ is also submodular. A function $f$ is submodular if for any two sets $A$ and $B$ in its domain, $f(A) + f(B) \geq f(A \cup B) + f(A \cap B)$.

    Let $A_{\mathcal{M}} \subseteq \mathcal{M}$ and $B_{\mathcal{M}} \subseteq \mathcal{M}$ be two sets of macro-facets. We must show:
    \[
    U'(A_{\mathcal{M}}) + U'(B_{\mathcal{M}}) \geq U'(A_{\mathcal{M}} \cup B_{\mathcal{M}}) + U'(A_{\mathcal{M}} \cap B_{\mathcal{M}}).
    \]
    Let $S_A = \operatorname{Exp}(A_{\mathcal{M}})$ and $S_B = \operatorname{Exp}(B_{\mathcal{M}})$ be the corresponding sets of raw facets. By definition, the inequality we need to prove is:
    \[
    U(S_A) + U(S_B) \geq U(\operatorname{Exp}(A_{\mathcal{M}} \cup B_{\mathcal{M}})) + U(\operatorname{Exp}(A_{\mathcal{M}} \cap B_{\mathcal{M}})).
    \]
    The expansion of a union of macro-facet sets is the union of their expansions: $\operatorname{Exp}(A_{\mathcal{M}} \cup B_{\mathcal{M}}) = S_A \cup S_B$. So the inequality becomes:
    \[
    U(S_A) + U(S_B) \geq U(S_A \cup S_B) + U(\operatorname{Exp}(A_{\mathcal{M}} \cap B_{\mathcal{M}})).
    \]
    Since $U$ is submodular by assumption, we know that:
    \[
    U(S_A) + U(S_B) \geq U(S_A \cup S_B) + U(S_A \cap S_B).
    \]
    Comparing these two inequalities, our proof is complete if we can show that $U(S_A \cap S_B) \geq U(\operatorname{Exp}(A_{\mathcal{M}} \cap B_{\mathcal{M}}))$. Since $U$ is monotone, this inequality holds if we can prove the underlying set relationship:
    \[
    S_A \cap S_B \supseteq \operatorname{Exp}(A_{\mathcal{M}} \cap B_{\mathcal{M}}).
    \]
    To prove this, let $x$ be an arbitrary element in $\operatorname{Exp}(A_{\mathcal{M}} \cap B_{\mathcal{M}})$. By definition, $x \in \operatorname{cl}_G(m)$ for some macro-facet $m \in A_{\mathcal{M}} \cap B_{\mathcal{M}}$.
    Since $m \in A_{\mathcal{M}}$, its closure $\operatorname{cl}_G(m)$ is a subset of $\operatorname{Exp}(A_{\mathcal{M}}) = S_A$.
    Similarly, since $m \in B_{\mathcal{M}}$, its closure $\operatorname{cl}_G(m)$ is a subset of $\operatorname{Exp}(B_{\mathcal{M}}) = S_B$.
    Therefore, $\operatorname{cl}_G(m) \subseteq S_A \cap S_B$. As $x \in \operatorname{cl}_G(m)$, it follows that $x \in S_A \cap S_B$. Since $x$ was arbitrary, the subset relationship holds.

    This completes the proof that $U'$ is submodular.
\end{proof}

\begin{proposition}[Preservation of Submodularity Ratio]
    If the utility function $U$ over the raw facets $\mathcal{C}_u$ is monotone and has a submodularity ratio of $\gamma \in [0, 1]$, then the utility function $U'$ over the macro-facets $\mathcal{M}$ is also monotone and has a submodularity ratio of at least $\gamma$.
\end{proposition}

\begin{proof}
    Monotonicity of $U'$ was established in the previous proposition. We now prove the preservation of the submodularity ratio.

    Let $A_{\mathcal{M}} \subseteq B_{\mathcal{M}} \subseteq \mathcal{M}$ be two sets of macro-facets, and let $X_{\mathcal{M}} \subseteq \mathcal{M} \setminus B_{\mathcal{M}}$ be another set of macro-facets. We must show that the marginal gain of adding $X_{\mathcal{M}}$ satisfies the diminishing returns property for $U'$:
    \[
    \Delta'(X_{\mathcal{M}} \mid A_{\mathcal{M}}) \geq \gamma \cdot \Delta'(X_{\mathcal{M}} \mid B_{\mathcal{M}}).
    \]
    By definition of $U'$, this is equivalent to:
    \[
    U'(A_{\mathcal{M}} \cup X_{\mathcal{M}}) - U'(A_{\mathcal{M}}) \geq \gamma \cdot (U'(B_{\mathcal{M}} \cup X_{\mathcal{M}}) - U'(B_{\mathcal{M}})).
    \]
    Let's define the corresponding expanded sets of raw facets:
    \begin{itemize}
        \item $S_A = \operatorname{Exp}(A_{\mathcal{M}})$
        \item $S_B = \operatorname{Exp}(B_{\mathcal{M}})$
        \item $S_X = \operatorname{Exp}(X_{\mathcal{M}})$
    \end{itemize}
    Since $A_{\mathcal{M}} \subseteq B_{\mathcal{M}}$, we know $S_A \subseteq S_B$. The inequality in terms of the original utility function $U$ is:
    \[
    U(S_A \cup S_X) - U(S_A) \geq \gamma \cdot (U(S_B \cup S_X) - U(S_B)).
    \]
    This is the definition of marginal gain for a set, so we can write it as:
    \[
    \Delta(S_X \mid S_A) \geq \gamma \cdot \Delta(S_X \mid S_B).
    \]
    The marginal gain of adding the macro-facets $X_{\mathcal{M}}$ to $A_{\mathcal{M}}$ can be written in terms of their raw facet expansions as $\Delta(S_X \mid S_A) = \Delta(S_X \setminus S_A \mid S_A)$.

Since $S_A \subseteq S_B$, we can decompose the set of new raw facets $S_X \setminus S_A$ into two disjoint sets:
\begin{itemize}
    \item $Z \triangleq S_X \setminus S_B$
    \item $W \triangleq S_X \cap (S_B \setminus S_A)$
\end{itemize}
such that $S_X \setminus S_A = Z \cup W$ and $Z \cap W = \emptyset$.

We can now establish a chain of inequalities:

\begingroup
\begin{align}
\Delta(S_X \mid S_A)
 &= \Delta(S_X \!\setminus\! S_A \mid S_A) \nonumber\\
\shortintertext{(Lemma~\ref{lem:marginal_superset})}
 &= \Delta(Z \cup W \mid S_A) \nonumber\\
\shortintertext{(decomposition)}
 &= U(S_A \cup Z \cup W) - U(S_A) \nonumber\\
\shortintertext{(definition)}
 &= [U(S_A \cup Z \cup W)-U(S_A \cup Z)] \nonumber\\
 &\hspace{1cm} + [U(S_A \cup Z)-U(S_A)] \nonumber\\
\shortintertext{(identity)}
 &= \Delta(W \mid S_A \cup Z) \nonumber\\
 &\hspace{1cm}+ \Delta(Z \mid S_A) \nonumber\\
\shortintertext{(marginal gain)}
 &\ge \Delta(Z \mid S_A) \nonumber\\
\shortintertext{(monotonicity)}
 &\ge \gamma\, \Delta(Z \mid S_B) \nonumber\\
\shortintertext{(submod. ratio, $S_A\!\subseteq\!S_B$)}
 &= \gamma\, \Delta(S_X\!\setminus\! S_B \mid S_B) \nonumber\\
\shortintertext{(def.\ of $Z$)}
 &= \gamma\, \Delta(S_X \mid S_B). \nonumber\\
\shortintertext{(Lemma~\ref{lem:marginal_superset})}
\end{align}
\endgroup

The chain of inequalities establishes that $\Delta(S_X \mid S_A) \geq \gamma \cdot \Delta(S_X \mid S_B)$. By the definition of $U'$, this is equivalent to $\Delta'(X_{\mathcal{M}} \mid A_{\mathcal{M}}) \geq \gamma \cdot \Delta'(X_{\mathcal{M}} \mid B_{\mathcal{M}})$, which completes the proof.
\end{proof}

\noindent\textbf{Theorem 1 (restated).}
    The laminar system $(\mathcal{M}, \mathcal{I})$ defined above is a matroid.
\begin{proof}
We prove that the laminar system $(\mathcal{M}, \mathcal{I})$ satisfies the three matroid axioms.
\begin{enumerate}
\item \textbf{Axiom 1 (Empty Set):} The empty set $\emptyset$ is in $\mathcal{I}$ because $|\emptyset \cap A|=0$ for all $A \in \mathcal{F}$, and $q_A \geq 0$ by definition. Thus, $0 \leq q_A$ for all $A \in \mathcal{F}$.

\item \textbf{Axiom 2 (Downward Closure):} Let $S \in \mathcal{I}$ and $S' \subseteq S$. For any set $A \in \mathcal{F}$, we have $S' \cap A \subseteq S \cap A$. This implies $|S' \cap A| \leq |S \cap A|$. Since $S \in \mathcal{I}$, we know $|S \cap A| \leq q_A$. Combining these, we get $|S' \cap A| \leq q_A$, which means $S' \in \mathcal{I}$.

\item \textbf{Axiom 3 (Augmentation Property):} Let $A, B \in \mathcal{I}$ with $|A| < |B|$. We must show that there exists an element $x \in B \setminus A$ such that $A \cup \{x\} \in \mathcal{I}$.

\end{enumerate}

Assume for contradiction that for every element $x \in B \setminus A$, the set $A \cup \{x\}$ is not in $\mathcal{I}$. By the definition of independence, this means for each $x \in B \setminus A$, there exists a set $F_x \in \mathcal{F}$ such that $x \in F_x$ and the quota for $F_x$ is violated, i.e., $|(A \cup \{x\}) \cap F_x| > q_{F_x}$.
Since $A \in \mathcal{I}$, we know $|A \cap F_x| \leq q_{F_x}$. The only way the union can violate the quota is if $|A \cap F_x| = q_{F_x}$.

This is a known result for laminar matroids. The most direct proof relies on the fact that for any $A, B \in \mathcal{I}$ with $|A| < |B|$, there is an element $x \in B \setminus A$ such that for every $F \in \mathcal{F}$ with $x \in F$, we have $|A \cap F| < q_F$.
This implies $A \cup \{x\}$ is independent.
Let's prove this claim. Assume for contradiction that for all $x \in B \setminus A$, there exists an $F_x \in \mathcal{F}$ such that $x \in F_x$ and $|A \cap F_x| = q_{F_x}$.
Let $S_{B \setminus A} = \bigcup_{x \in B \setminus A} \{x\}$.

Let $\mathcal{T} = \{F \in \mathcal{F} \mid |A \cap F| = q_F\}$ be the family of sets in $\mathcal{F}$ that are "tight" with respect to the set $A$. By our assumption, for each $x \in B \setminus A$, the collection of sets in $\mathcal{T}$ containing $x$ is non-empty.

Now, let $\mathcal{T}_{\min} \subseteq \mathcal{T}$ be the family of minimal sets in $\mathcal{T}$ under set inclusion. 

\textbf{Claim 1} (Pairwise Disjoint Sets). \textit{The sets in }$\mathcal{T}_{\min}$ \textit{are pairwise disjoint.  }

\begin{proof}[Proof of Claim 1]
Let $T_1, T_2$ be two distinct sets in $\mathcal{T}_{\min}$. If they were not disjoint ($T_1 \cap T_2 \neq \emptyset$), then by the laminar property of $\mathcal{F}$, one must be a proper subset of the other (as they are not equal). This would contradict the assumption that both $T_1$ and $T_2$ are minimal sets in $\mathcal{T}$. Therefore, any two distinct sets in $\mathcal{T}_{\min}$ must be disjoint.
\end{proof}
For any element $x \in B \setminus A$, there exists at least one set $F \in \mathcal{T}$ containing it. This means there must be a minimal such set $T \subseteq F$ in $\mathcal{T}_{\min}$ that also contains $x$. Thus, every element of $B \setminus A$ is contained in some set from the disjoint family $\mathcal{T}_{\min}$.

Let $\{T_1, \ldots, T_k\}$ be the members of $\mathcal{T}_{\min}$ that contain at least one element from $B \setminus A$. This is a pairwise disjoint family of sets, and their union covers $B \setminus A$.

For each $T_i$, there exists at least one element $x \in B \setminus A$ such that $x \in T_i$, and $|A \cap T_i| = q_{T_i}$.
Since $B \in \mathcal{I}$, we have $|B \cap T_i| \leq q_{T_i}$.
Since each $T_i$ is in $\mathcal{T}$, it is "tight" with respect to $A$, which means $|A \cap T_i| = q_{T_i}$ by definition. Furthermore, since $B$ is an independent set in $\mathcal{I}$, it must satisfy all quota constraints, so $|B \cap T_i| \leq q_{T_i}$. Combining these two facts, we have $|A \cap T_i| = q_{T_i} \geq |B \cap T_i|$.

To see why this implies $|(A \setminus B) \cap T_i| \geq |(B \setminus A) \cap T_i|$, we decompose the sets $A \cap T_i$ and $B \cap T_i$ based on their intersection with $A$ and $B$.
\begin{align}
|A \cap T_i|
&= |(A \cap T_i) \cap B| + |(A \cap T_i) \setminus B| \nonumber\\
&= |A \cap B \cap T_i| + |(A \setminus B) \cap T_i|, \label{eq:a}\\[4pt]
|B \cap T_i|
&= |(B \cap T_i) \cap A| + |(B \cap T_i) \setminus A| \nonumber\\
&= |A \cap B \cap T_i| + |(B \setminus A) \cap T_i|.
\end{align}

Since we know $|A \cap T_i| \geq |B \cap T_i|$, we can substitute these expressions:
\[
|A \cap B \cap T_i| + |(A \setminus B) \cap T_i| \geq |A \cap B \cap T_i| + |(B \setminus A) \cap T_i|.
\]
Subtracting the common term $|A \cap B \cap T_i|$ from both sides yields the desired inequality, $|(A \setminus B) \cap T_i| \geq |(B \setminus A) \cap T_i|$.
By summing this inequality over all disjoint sets $T_i$ that cover $B \setminus A$, we get:
\[
\sum_{i=1}^k |(A \setminus B) \cap T_i| \geq \sum_{i=1}^k |(B \setminus A) \cap T_i|.
\]
Since the sets $T_i$ are disjoint, the sum on the right side is $|(B \setminus A) \cap (\bigcup_{i=1}^k T_i)|$. Because the family $\{T_i\}$ covers all of $B \setminus A$, this is equal to $|B \setminus A|$.
The sum on the left side is $|(A \setminus B) \cap (\bigcup_{i=1}^k T_i)|$, which is the size of a subset of $A \setminus B$. Therefore, we have the following chain of inequalities:
\begin{align}
|A \setminus B|
&\ge |(A \setminus B) \cap (\bigcup_{i=1}^k T_i)| \nonumber\\
&= \sum_{i=1}^k |(A \setminus B) \cap T_i| \label{eq:asymdiff-chain}\\
&\ge \sum_{i=1}^k |(B \setminus A) \cap T_i| \nonumber\\
&= |B \setminus A|. \nonumber
\end{align}

This means $|A \setminus B| \geq |B \setminus A|$, which contradicts our initial assumption that $|A|<|B|$, as that implies $|A \setminus B| < |B \setminus A|$.
Therefore, our original assumption must be false. There must exist an element $x \in B \setminus A$ such that for all $F \in \mathcal{F}$ with $x \in F$, we have $|A \cap F| < q_F$. This element $x$ can be added to $A$ to form an independent set $A \cup \{x\}$, since it does not violate any quota.

Since all three axioms are satisfied, the laminar system $(\mathcal{M}, \mathcal{I})$ is a matroid.
\end{proof}

\begin{corollary}[Partition Matroid]
\label{cor:partition-matroid}
The system $(\mathcal{M}, \mathcal{I})$ defined by a partition of the ground set $\mathcal{M}=\biguplus_{t \in T} \mathcal{M}_t$ with per-partition quotas $q_t$ is a matroid.
\end{corollary}

\begin{proof}
A partition matroid is a special case of a laminar matroid. We can define a laminar family $\mathcal{F} = \{\mathcal{M}_t \mid t \in T\}$ over the partitioned ground set. This family is laminar because for any two distinct sets $\mathcal{M}_{t_1}, \mathcal{M}_{t_2} \in \mathcal{F}$, their intersection is the empty set, $\mathcal{M}_{t_1} \cap \mathcal{M}_{t_2} = \emptyset$.

The independence condition for a set $S \subseteq \mathcal{M}$ in this laminar system is $|S \cap \mathcal{M}_t| \leq q_t$ for all $t \in T$, which is precisely the definition of an independent set in a partition matroid. Since we have proven in Theorem~\ref{thm:laminar-matroid} that any such laminar system is a matroid, it follows directly that the partition system is also a matroid.
\end{proof}

\end{document}